\documentclass[sigconf]{acmart}
\AtBeginDocument{%
  }

\setcopyright{acmlicensed}
\copyrightyear{2018}
\acmYear{2018}
\acmDOI{XXXXXXX.XXXXXXX}
\acmConference[Conference acronym 'XX]{Make sure to enter the correct
  conference title from your rights confirmation email}{June 03--05,
  2018}{Woodstock, NY}
\acmISBN{978-1-4503-XXXX-X/2018/06}
\usepackage{pdflscape}
\usepackage{multirow}
\usepackage{adjustbox}
\usepackage{tcolorbox}
\usepackage{verbatim}
\usepackage{enumitem}
\usepackage{float}
\usepackage{graphicx}
\usepackage{makecell}
\usepackage{mdframed}
\usepackage{subcaption}
\usepackage{hyperref}

\begin{document}

\title{C-TLSAN: Content-Enhanced Time-Aware Long- and Short-Term Attention Network for Personalized Recommendation}



\author{Siqi Liang}
\authornote{Both authors contributed equally to this research.}
\email{lsq950917@gmail.com}
\affiliation{%
 \institution{Purdue University}
 \city{West Lafayette}
 \state{Indiana}
 \country{USA}
}
\author{Yudi Zhang}
\authornotemark[1]
\email{yudiz@iastate.edu}
\affiliation{%
 \institution{Iowa State University}
 \city{Ames}
 \state{Iowa}
 \country{USA}
}

\author{Yubo Wang}
\email{wangyubo9398@gmail.com}
\affiliation{%
 \institution{Purdue University}
 \city{West Lafayette}
 \state{Indiana}
 \country{USA}
}


\begin{abstract}
Sequential recommender systems aim to model users’ evolving preferences by capturing patterns in their historical interactions. Recent advances in this area have leveraged deep neural networks and attention mechanisms to effectively represent sequential behaviors and time-sensitive interests. In this work, we propose C-TLSAN (Content-Enhanced Time-Aware Long- and Short-Term Attention Network), an extension of the TLSAN architecture that jointly models long- and short-term user preferences while incorporating semantic content associated with items—such as product descriptions.

C-TLSAN enriches the recommendation pipeline by embedding textual content linked to users’ historical interactions directly into both long-term and short-term attention layers. This allows the model to learn from both behavioral patterns and rich item content, enhancing user and item representations across temporal dimensions. By fusing sequential signals with textual semantics, our approach improves the expressiveness and personalization capacity of recommendation systems.

We conduct extensive experiments on large-scale Amazon datasets, benchmarking C-TLSAN against state-of-the-art baselines, including recent sequential recommenders based on Large Language Models (LLMs), which represent interaction history and predictions in text form. Empirical results demonstrate that C-TLSAN consistently outperforms strong baselines in next-item prediction tasks. Notably, it improves AUC by 1.66\%, Recall@10 by 93.99\%, and Precision@10 by 94.80\% on average over the best-performing baseline (TLSAN) across 10 Amazon product categories. These results highlight the value of integrating content-aware enhancements into temporal modeling frameworks for sequential recommendation. Our code is available at \url{https://github.com/booml247/cTLSAN}.
 
\end{abstract}



\received{20 February 2007}
\received[revised]{12 March 2009}
\received[accepted]{5 June 2009}

\maketitle

\section{Introduction}

With the growing demand for personalized user experiences, delivering timely and semantically rich recommendations has become a cornerstone of digital marketing. Traditional static models are being replaced by methods that capture the dynamic nature of user behavior\cite{Quadrana2018}. Sequential recommendation reflects this shift by modeling interactions as ordered sequences, uncovering temporal dependencies and behavioral patterns \cite{Fang2020, Hidasi2016}.
Deep learning has accelerated progress in this area, enabling models to learn complex user-item interactions and time-sensitive patterns \cite{Wang2019}. These methods have gained traction across domains such as e-commerce and streaming, where understanding the temporal flow of user interests enhances engagement and conversion \cite{Sun2019, Chen2018}.

Despite these advances, real-world deployment of sequential recommenders remains challenging. Practical systems must model both short- and long-term dependencies while incorporating diverse signals—such as user histories, item metadata (e.g., price, brand), and textual content (e.g., titles, descriptions). They must also scale efficiently, adapt to user changes, and support rapid experimentation.

Recently, Large Language Models (LLMs) have been applied to sequential recommendation by framing the task as language modeling—encoding user histories and predicting items using text prompts \cite{Qu2024, Li2024}. Though promising, these methods often struggle to capture fine-grained temporal patterns and structured behavioral signals.

To address these limitations, we propose C-TLSAN (Content-Aware Time-aware Long- and Short-term Attention Network), an extension of TLSAN \cite{ZHANG2021}, which integrates item-level textual content into temporal modeling. C-TLSAN enhances user and item representations by incorporating product descriptions and metadata into both long- and short-term attention layers, enabling more robust and personalized recommendations.
The main contributions of this work include: 
\begin{itemize}
\item We introduce C-TLSAN, a content-aware sequential recommendation model that unifies temporal attention with item-level textual content, improving the quality of user-item representations.
\item We benchmark C-TLSAN against state-of-the-art LLM-based recommendation methods, showing that our model consistently outperforms them in accuracy and robustness. We also highlight the limitations of LLM-based approaches in capturing structured and temporal patterns.
\item Extensive experiments on ten large-scale Amazon datasets demonstrate that C-TLSAN delivers significant improvements over competitive baselines across multiple metrics.
\end{itemize}

\section{Related Work}
\begin{figure*}[htbp]
    \centering
    \includegraphics[width=0.85\linewidth]{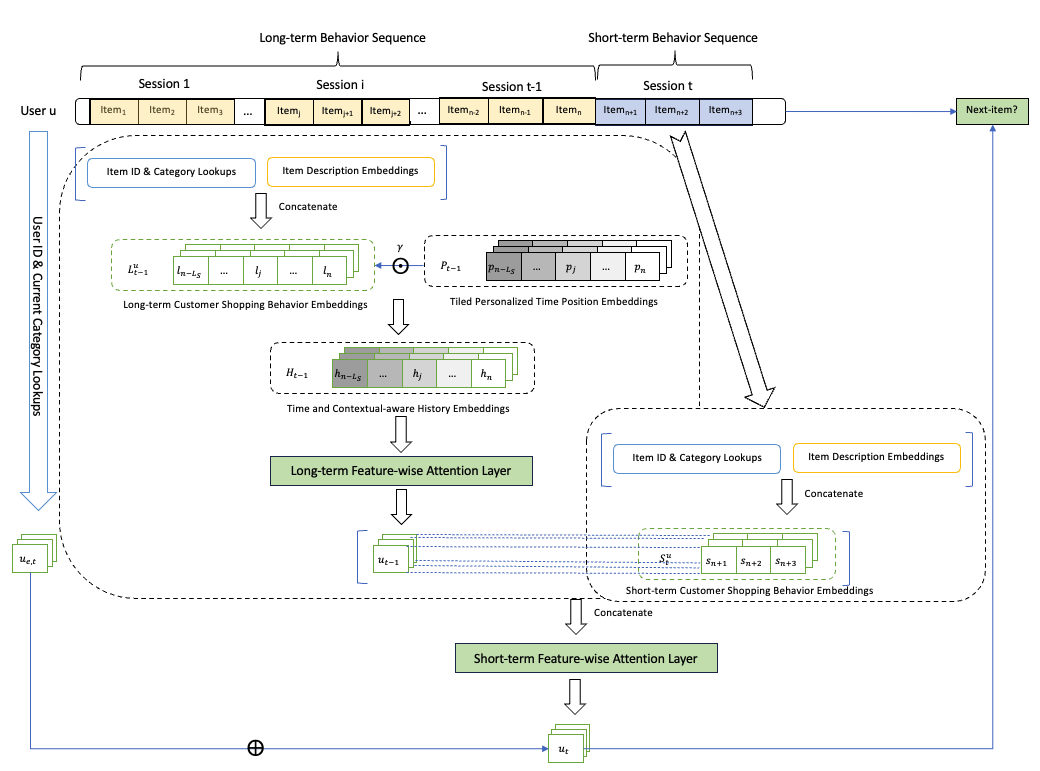}
    \caption{Model architecture of C-TLSAN.}
    \Description{A block diagram showing the C-TLSAN model architecture, including embedding layers, time-aware attention mechanisms for long-term and short-term user preferences, and the final prediction layer.}
    \label{fig:cTLSAN}
\end{figure*}

Sequential recommendation has attracted substantial research attention in recent years~\cite{ZHANG2021, Huang2018, Cao2020, Ying2018, Zhou2018}. SHAN~\cite{Ying2018} leverages an attention mechanism to model both long- and short-term user preferences, incorporating context through user embeddings. However, it overlooks the effect of time decay in long-term behavior sequences and fails to capture fine-grained user-item interactions across different feature dimensions.
Building on these limitations, \citet{ZHANG2021} proposed TLSAN, which introduces a dual-attention mechanism to independently model long-term and short-term user behaviors while explicitly incorporating temporal dynamics. Inspired by the Deep Interest Network (DIN)~\cite{zhou2018deep}, TLSAN effectively balances recent user interests with historically persistent preferences by applying personalized, time-aware attention mechanisms to both temporal scopes.

While transformer-based architectures dominate sequential recommendation, \citet{fan2024} introduced TiM4Rec, which employs a time-aware structured state space model. This novel approach offers an alternative to attention mechanisms by utilizing state space modeling to capture temporal dynamics and sequential dependencies in user behaviors.

The integration of LLMs in recommendation systems has emerged as a promising direction.~\citet{yang2024} proposed LRD (Sequential Recommendation with Latent Relations based on LLM), which leverages LLMs to discover and utilize implicit relationships between products, enriching the recommendation model with semantic understanding. \citet{Qu2024} presented a critical examination of pre-trained language models in sequential recommendation, demonstrating how LLM embeddings can be effectively used to initialize ID embeddings in traditional deep learning recommendation models. Their work bridges the gap between semantic-rich LLM representations and conventional recommendation architectures. Taking a more radical approach, \citet{Li2024} proposed CALRec, a purely LLM-based recommendation framework that utilizes contrastive alignment of generative LLMs for sequential recommendation tasks.

\section{Method}

\subsection{Problem Formulation}
Let 
$
\mathcal{U} = \left\{u_1, u_2, \ldots, u_M\right\}
$
, 
$
\mathcal{I} = \left\{j_1, j_2, \ldots, j_N\right\}
$
and 
$
\mathcal{C} = \left\{c_1, c_2, \ldots, c_K\right\}
$
denote the sets of users, items and item categories, respectively. For each user $u \in \mathcal{U}$, we define their interaction history up to time $t$ as
$
\mathcal{L}_t^u = \left\{\mathcal{S}_1^u, \mathcal{S}_2^u, \ldots, \mathcal{S}_t^u\right\}
$,
where each $\mathcal{S}_i^u\subseteq \mathcal{I}(i \in[1,t])$ represents a session containing items the user interacted with at time $i$. 

Following the architecture of TLSAN, we divide the interaction history into two parts: the long-term behavior sequence $\mathcal{L}_{t-1}^u$, capturing user preferences across past sessions, and the short-term behavior sequence $\mathcal{S}_t^u$, representing recent session-level interactions, where $t$ is the current time.

The goal is to accurately predict the next item a user is likely to interact with, based on their historical behavior, and recommend the most relevant items accordingly.

\subsection{Fusing Content Information}
Baseline models typically represent users, items, and categories using discrete ID embeddings. While effective in capturing collaborative signals, such approaches overlook valuable item-level content information—such as product descriptions and metadata—which can provide richer semantic cues about item characteristics and user intent.

To address this limitation, we enhance the model by integrating content-based item representations. Specifically, we define the item content set as $\mathcal{M} = {m_1, m_2, \dots, m_N}$, where $m_i$ corresponds to the content features (e.g., title, description) of item $i \in \mathcal{I}$. These content features are embedded into dense semantic vectors and fused with traditional ID-based embeddings, enabling the model to capture more nuanced item semantics and improve its ability to generate relevant recommendations.

The remaining architecture follows TLSAN~\cite{ZHANG2021}. During training, we use all user interaction histories in the training set to learn user-item interaction patterns. The model operates in a binary classification setting, predicting whether a user will engage with a candidate item in the test set. Figure~\ref{fig:cTLSAN} illustrates the architecture of C-TLSAN. User behavior histories are segmented into long-term and short-term sessions based on timestamps, and a random item from the short-term session is selected as the prediction target. The model comprises three main components:

(a) Dynamic User Category Extraction: From the long-term behavior sequence and category embeddings, a dynamic user category is derived to obtain a user-specific category representation $u_{e,t}$.

(b) Long-Term Preference Extraction: For the long-term session, item ID and category embeddings are concatenated with item content embeddings to form the long-term behavior representation $\mathcal{L}{t-1}^u$. Personalized time position embeddings $\mathcal{P}{t-1}$ are added to model user-specific temporal dynamics, yielding temporally and semantically enriched historical embeddings $\mathcal{H}{t-1}$. Feature-wise attention is applied to extract the user’s long-term preferences, denoted as $u{t-1}$.

(c) Short-Term Preference Extraction: Similarly, item ID and category embeddings are combined with content embeddings for the short-term session to produce $\mathcal{S}{t}^u$. This is concatenated with $u{t-1}$, and feature-wise attention is applied to compute the unified user representation $u_{t}$.

For final prediction, the user category representation $u_{e,t}$ and short-term preference $u_{t}$ are summed to form a comprehensive user vector. A sigmoid cross-entropy loss is used, where the positive sample is the ground-truth next item and negative samples are randomly drawn. This loss function encourages the model to prioritize relevant items while suppressing irrelevant ones.
\subsection{LLM Based Sequential Recommendation}

\begin{figure}[h]
\vspace{-0.5cm}
\begin{tcolorbox}[
    width=\linewidth,       
    boxsep=1pt,                
    left=1pt, right=1pt,       
    top=2pt, bottom=2pt,       
    arc=4pt,                   
]
\normalsize
\begin{quote}
You are a smart recommendation assistant.

Here is the customer's recent browsing history:
\begin{verbatim}
1. [Product description 1]
2. [Product description 2]
3. [Product description 3]
...
\end{verbatim}

Now, we want to recommend \textbf{one} of the following two items:

\begin{verbatim}
[Product ID 1]: [Product description]
[Product ID 2]: [Product description]
\end{verbatim}

Based on the customer's interest, which item would you recommend? Respond in the format:

Recommendation: [Product ID 1 or Product ID 2]\\
Reason: [your explanation]
\end{quote}
\end{tcolorbox}
\vspace{-0.3cm}
\caption{Example prompt for sequential product recommendation task.}
\Description{Example prompt for sequential product recommendation task.}
\label{fig:recommendation_prompt_example}
\end{figure}
Inspired by recent advances in sequential recommendation using LLMs~\cite{Zheng2024}, we explore a simplified yet effective approach that leverages prompt engineering to instruct LLMs in next-item prediction tasks. The core idea is to reframe the sequential recommendation problem as a natural language understanding task, where a user's interaction history and candidate items are expressed in text form and passed as input to a pretrained LLM (LLAMA 3.3 70B~\cite{touvron2024llama3}).

Figure~\ref{fig:recommendation_prompt_example} illustrates an example prompt template used in our experiments. The prompt presents the model with a user's browsing history encoded as a list of product descriptions, followed by two candidate items. The LLM is then asked to choose the most appropriate next item and explain its reasoning in natural language.

In our experiments, we compare this prompt-based LLM recommendation approach against C-TLSAN and other neural baselines. The results demonstrate that while LLMs exhibit promising performance in some scenarios, C-TLSAN still outperforms them consistently across most datasets, highlighting the benefits of dedicated temporal modeling and the integration of rich item content information.

\begin{table*}[htbp]
\caption{AUC on the public Amazon product datasets. Bold font and underlined value indicate
the optimal result and the suboptimal result, separately.}
\begin{tabular}{l|ccccccc}
\hline
Datasets                      & LLM Recommender & CSAN  & ATRank & Bi-LSTM & PACA  & TLSAN & C-TLSAN \\ \hline
CDs and Vinyl               & 0.827           & 0.813 & 0.889  & 0.881   & 0.801 & \textbf{0.942} & \underline{0.938}  \\
Clothing Shoes and Jewelry & 0.639           & 0.761 & 0.663  & 0.668   & 0.798 & \underline{0.927} & \textbf{0.938}  \\
Digital-Music                 & 0.808           & 0.729 & 0.825  & 0.792   & 0.963 & \underline{0.972} & \textbf{0.974}  \\
Office Products              & 0.516           & 0.824 & 0.921  & 0.856   & 0.910 & \underline{0.969} & \textbf{0.976}  \\
Movies and TV            & 0.813           & 0.797 & 0.860  & 0.824   & 0.806 & \underline{0.879} & \textbf{0.909}  \\
Beauty                        & 0.618           & 0.727 & 0.806  & 0.773   & 0.859 & \underline{0.925} & \textbf{0.947}  \\
Home and Kitchen            & 0.592           & 0.702 & 0.736  & 0.684   & 0.788 & \underline{0.865} & \textbf{0.895}  \\
Video Games                  & 0.587           & 0.807 & 0.870  & 0.820   & \underline{0.917} & 0.914 & \textbf{0.933}  \\
Toys and Games              & 0.678           & 0.812 & 0.829  & 0.775   & 0.861 & \underline{0.922} & \textbf{0.936}  \\
Electronics                   & 0.587           & 0.811 & 0.841  & 0.811   & 0.835 & \underline{0.894} & \textbf{0.913}  \\ \hline
\end{tabular}
\label{table: AUC}
\end{table*}

\section{Results}

\subsection{Metrics Comparison}


To comprehensively evaluate the effectiveness of C-TLSAN across various domains, we conduct experiments on the publicly available datasets from Amazon\footnote{\href{http://jmcauley.ucsd.edu/data/amazon/}{http://jmcauley.ucsd.edu/data/amazon/}}, a large-scale collection of user-item interaction records across a wide range of product categories. The dataset contains detailed information on users, items, reviews, and product metadata, making it a widely adopted benchmark for recommendation research. The datasets are preprocessed following two steps:
\begin{itemize}
    \item Filtering: Users with fewer than 10 interactions and items with fewer than 8 interactions are removed to ensure data quality.
    \item Behavior Range: Users with between 5 and 90 transactions are retained to capture both long-term and short-term behavior patterns within a recent three-month period.
\end{itemize}
User behavior records are segmented into ordered sessions based on daily interactions. For training, the target (next) item is randomly selected from the newest session at time step $t$.
If the session contains only one item, the first item from the next session (time $t+1$) is chosen instead.
For models explicitly handling long- and short-term user behaviors (e.g., TLSAN\cite{ZHANG2021}):
The newest session (excluding the target item) is treated as the short-term context, while earlier sessions (1 to $t-1$) form the long-term history.
For other baseline models (ATRank\cite{Zhou2018}, Bi-LSTM\cite{jozefowicz2015empirical}, PACA\cite{Cao2020}, CSAN\cite{Huang2018}, LLM), all sessions before time $t$ are used as historical interactions.
The item following the newest session is selected as the test item to evaluate the models. 
The pre-trained embedding model used in C-TLSAN is \textit{all-MiniLM-L6-v2}.
To ensure the fairness and comparability of the experiments, we keep the common parameters in each model the same and the unique parameters the optimal.
All the experiments in this paper are implemented with Python 3.11.11 and Tensorflow 2.17.0, and run on a server with 192 vCPUs (AMD EPYC 2nd Gen) and 8 $\times$ NVIDIA A10G (192 GB total GPU memory).

\begin{figure*}[htbp]
  \centering
  \subcaptionbox{Recall@K\label{fig:recall}}[0.48\linewidth]{
    \includegraphics[width=\linewidth]{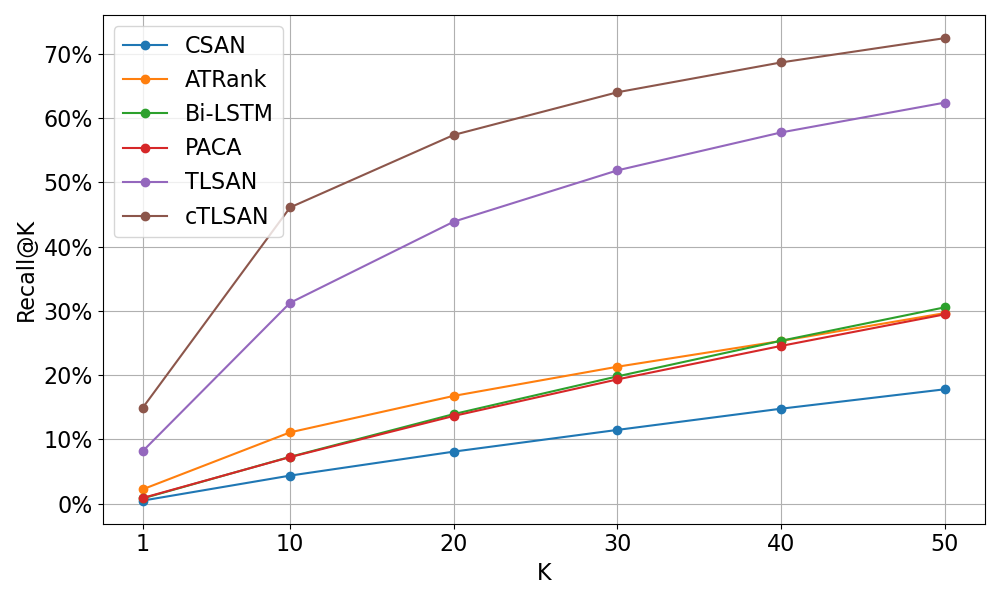}
  }
  \hfill
  \subcaptionbox{Precision@K\label{fig:precision}}[0.48\linewidth]{
    \includegraphics[width=\linewidth]{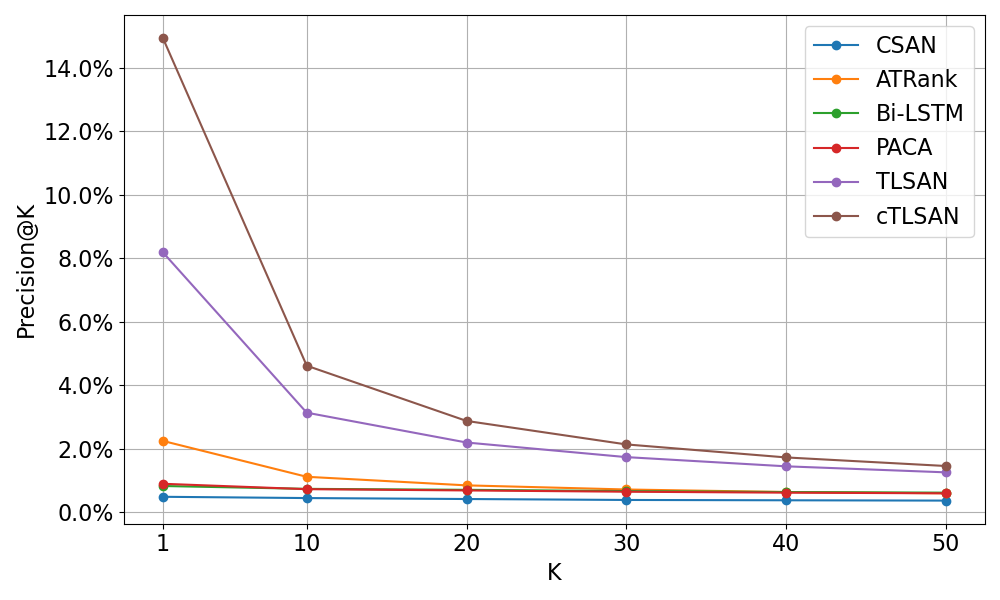}
  }
  \caption{Recall@K and Precision@K on Amazon Office Products dataset.}
  \Description{Recall@K and Precision@K on Amazon Office Products dataset.}
  \label{fig:recall_and_precision}
\end{figure*}

Table \ref{table: AUC}, \ref{table: Precision&Recall} and Figure \ref{fig:recall_and_precision} summarize the performance comparisons. The proposed C-TLSAN is the best performer while its predecessor TLSAN performs the second best. Across all datasets, cTLSAN either matches or surpasses TLSAN, demonstrating the added value of integrating textual item information. On average, cTLSAN improves over TLSAN by 1.66\% in AUC, with notable gains observed in categories such as Office Products (from 0.969 to 0.976), Beauty (from 0.925 to 0.947), and Home and Kitchen (from 0.865 to 0.895). Besides, it enhances Recall@10 by 93.99\% and Precision@10 by 94.80\% on average over the best-performing baseline TLSAN across Amazon product datasets. These results validate the effectiveness of fusing content semantics with time-aware attention for capturing richer user-item relationships.
\begin{table}[htbp]
\caption{Recall@10 and Precision@10 on the public Amazon product datasets.}
\begin{tabular}{l|cccc}
\hline
                   & \multicolumn{2}{c}{Recall @10} & \multicolumn{2}{c}{Precision @10} \\ \hline
Dataset            & TLSAN          & cTLSAN        & TLSAN           & cTLSAN          \\
CDs\_and\_Vinyl    & 4.12\%         & 8.57\%        & 0.41\%          & 0.86\%          \\
Digital-Music      & 16.38\%        & 24.87\%       & 1.64\%          & 2.49\%          \\
Office\_Products   & 31.25\%        & 46.09\%       & 3.13\%          & 4.61\%          \\
Movies\_and\_TV\_5 & 1.64\%         & 6.04\%        & 0.16\%          & 0.60\%          \\
Beauty             & 11.27\%        & 18.54\%       & 1.13\%          & 1.85\%          \\
Home\_and\_Kitchen & 7.27\%         & 12.90\%       & 0.73\%          & 1.29\%          \\
Video\_Games       & 7.88\%         & 12.28\%       & 0.79\%          & 1.23\%          \\
Toys\_and\_Games   & 10.53\%        & 20.24\%       & 1.05\%          & 2.02\%          \\
Electronics        & 4.72\%         & 8.51\%        & 0.47\%          & 0.85\%          \\ \hline
\end{tabular}
\label{table: Precision&Recall}
\end{table}

In light of the recent surge in interest surrounding LLMs, we also compare C-TLSAN with a prompt-engineered LLM-based recommender. While the LLM recommender demonstrates competitive performance in a few domains—such as CDs and Vinyl and Movies and TV—it consistently underperforms compared to C-TLSAN across all evaluated datasets. The average AUC gap is considerable, underscoring the limitations of prompt-based LLMs in modeling fine-grained temporal dependencies and structured behavioral patterns. For instance, in the Office Products category, C-TLSAN achieves an AUC of 0.976, significantly outperforming the LLM-based method, which scores just 0.516. These results highlight the strengths of specialized architectures like C-TLSAN, which are explicitly designed to integrate both textual semantics and time-sensitive behavioral signals in sequential recommendation scenarios. Methodologically, the performance gap stems from several key limitations of the LLM-based approach:
\begin{itemize}
    \item LLMs treat user behavior history as plain text, with no inherent understanding of time intervals, session boundaries, or recency bias—factors that are central to effective sequential recommendation.
    \item While LLMs excel in capturing semantic relationships from free text, they struggle to model structured signals like category hierarchies, frequency patterns, and multi-session dependencies without additional scaffolding or feature engineering.
    \item The effectiveness of LLM-based recommendation is highly dependent on prompt design. Small changes in prompt format or phrasing can lead to significant performance variation, making the model harder to optimize and less reliable in production settings.
    \item LLMs require large-scale inference resources and can be prohibitively expensive for real-time applications, whereas C-TLSAN is comparatively lightweight and designed for scalable deployment.
\end{itemize}

In contrast, cTLSAN addresses these challenges directly. By incorporating both long- and short-term user behavior through personalized attention mechanisms and fusing textual information (e.g., item descriptions) into the representation space, it captures both semantic and sequential dynamics in a unified and interpretable framework.

\subsection{Limitations of LLMs in Content-Aware Recommendation}

A key limitation of using LLMs for recommendation is their tendency to rely on surface-level lexical similarity rather than deeper user intent or behavioral context.

Consider a user whose browsing history includes: (1) Avery 5267 bright white laser printer labels, (2) Pilot B2P recycled pens, (3) GP Ink Jet and Laser Paper, (4) a dual-tip eraser, and (5) Pilot’s Metropolitan Collection pens. This sequence reflects a clear behavioral pattern centered on printing, writing tools, and workspace optimization. However, the LLM recommended a \textbf{Post-it Super Sticky Color-Coding Label} instead of a more functionally relevant product, such as a \textbf{desktop printer stand with casters}. This misstep was driven by overemphasis on the keyword “labels” rather than an understanding of the user's broader task-oriented goal: optimizing a high-usage printing environment.

A similar issue arose in a scenario where the user's activity involved packing and organization tasks—browsing items such as a desktop file case, wardrobe moving boxes, and heavy-duty sealing tape. Despite this consistent behavioral context, the LLM recommended \textbf{binder dividers} instead of a more practical choice like \textbf{non-toxic, multicolor chalk}, which could be used for labeling or categorizing during the packing process. The model likely over-weighted weak textual associations (e.g., between “tabs” and “folders”) while missing the task-relevant semantics underlying the user’s actions.

In both cases, a deep learning model trained on behavioral sequences—such as co-clicks, purchases, and session-level intent—could more effectively infer the user's underlying goal and produce semantically coherent recommendations. This highlights the importance of leveraging structured behavioral context and temporal signals alongside textual understanding in recommendation systems.

\section{Conclusion}
In this paper, we introduced C-TLSAN, a content-aware, time-sensitive sequential recommendation model designed to capture both long- and short-term user preferences while integrating rich semantic item information such as product descriptions and metadata. By enhancing traditional ID-based embeddings with item-level content features and modeling user behavior through temporal attention mechanisms, C-TLSAN provides a more nuanced understanding of user intent.

Through comprehensive experiments on ten large-scale Amazon datasets, we demonstrated that C-TLSAN consistently surpasses state-of-the-art baselines, including recent approaches based on Large Language Models (LLMs). While LLMs offer impressive generalization and zero-shot capabilities, our findings highlight their limitations in capturing fine-grained temporal patterns and structured behavioral context—factors that are essential for high-quality personalized recommendations.

Looking forward, we see promising research directions in developing hybrid models that combine the generative reasoning and flexibility of LLMs with the structure-aware, behavior-driven strengths of deep recommendation networks. Such models could dynamically adapt to user context, incorporate multimodal content (e.g., images, user reviews), and enable more explainable and controllable recommendations. 

\bibliography{ref}
\bibliographystyle{ACM-Reference-Format}
\end{document}